\documentclass[11pt,a4paper]{article}
\usepackage[hyperref]{acl2019}
\usepackage{times}
\usepackage{latexsym}

\usepackage{url}
\usepackage{CJKutf8}  
\usepackage{pinyin} 
\usepackage{amsmath}
\usepackage{multirow}
\usepackage{graphicx}

\aclfinalcopy 

\title{A Simple Dual-decoder Model for Generating Response with Sentiment}

\author{Wu Xiuyu \\
  Peking University \\
  \texttt{xiuyu\_wu@pku.edu.cn} \\\And
  Wu Yunfang \\
  Peking University \\
  \texttt{wuyf@pku.edu.cn} \\}

\date{}

\begin{document}
\maketitle
\begin{abstract}
How to generate human like response is one of the most challenging tasks for artificial intelligence. In a real application, after reading the same post different people might write responses with positive or negative sentiment according to their own experiences and attitudes. To simulate this procedure, we propose a simple but effective dual-decoder model to generate response with a particular sentiment, by connecting two sentiment decoders to one encoder. To support this model training, we construct a new conversation dataset with the form of $(post, resp1, resp2)$ where two responses contain opposite sentiment. Experiment results show that our dual-decoder model can generate diverse responses with target sentiment, which obtains significant performance gain in sentiment accuracy and word diversity over the traditional single-decoder model. We will make our data and code publicly available for further study. 
\end{abstract}

\section{Introduction}
\label{sect:introduction}
For a long time, building an excellent conversation system which can generate meaningful and natural responses has attracted much attention in both academia and industry. Emotional factor is of great importance for response generation, because response with proper sentiment could motivate people to go on talking \cite{prendinger2005empathic} while response with wrong sentiment might end the conversation \cite{martinovski2003breakdown}. 

Recently, there are some works that deal with sentiment or emotional response generation. \cite{zhou2017mojitalk} employ Conditional Variational Autoencoder to train an emotional response generation model; \cite{zhou2018emotional} utilize emotion category embeddings in the decoding process; \cite{li2018syntactically} firstly generate two keywords and then use a bidirectional-asynchronous decoder to generate sentences. \cite{shi2018sentiment} employ reinforcement learning for training end-to-end dialog systems. Some other works try to avoid generating short, repetitive or meaningless responses \cite{zhang2018-avoid, pandey2018exemplar}. These existing work adopt a single-decoder architecture to deal with different sentiment and emotions. 

There are some approaches trying to improve the RNN model by using multi decoders. Deliberations network has two levels of decoders which are used in turn \cite{xia2017deliberation}. \citeauthor{geng2018adaptive} propose a model called adaptive multi-pass decoder for machine translation, where the result is passed thought several decoders in turn \cite{geng2018adaptive}. Moreover, two decoders are also used in the bidirectional-asynchronous decoder architecture, which firstly using some keywords to generate the forward "half" sentence and then generate the backward "half" sentence \cite{zhou2018emotional, mou2016sequence}. As can be seen, these approaches are all in complicated model structures. 

Different from the previous work, this paper proposes a simple but effective dual-decoder framework to generate different sentiment responses. According to our statistics on the short text conversation (STC) dataset \cite{Sina2013}, there are 194,305 posts corresponding with 5,648,129 responses, where 58.3\% of them (113,297 posts) have at least one pair of response in which one is positive while the other is negative. Figure 1 gives an example extracted from the real data. Unfortunately, there's little research to deal with generating different responses with target sentiment towards one post. 

\begin{figure}
\centering
\includegraphics[scale=0.44]{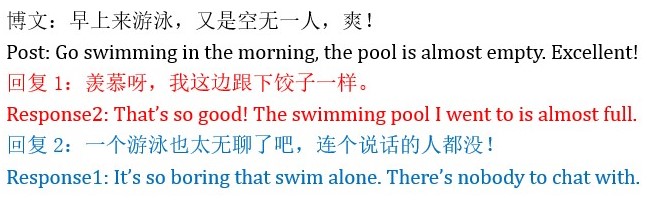}
\caption{\label{data-exp} Example of two responses with different sentiment towards one post in the STC data.}
\end{figure}

To model this data distribution, we reconstruct the existing conversation dataset. For one post, we collect $(resp1, resp2)$ pairs in which one response is positive and the other is negative. Based on the new dataset, we propose a new dual-decoder architecture to generate different sentiment responses with respect to one post. We connect two decoders to one encoder, which simulates that after reading the same post different people might have different thought and then generate different responses. The experiment results show that our model has excellent ability to generate responses with target sentiment given the same post, which greatly improves sentiment accuracy and word diversity over the traditional single-decoder model.

\section{Dataset Construction}
\label{sect:dataset collection}
We use the public STC dataset\footnote{http://ntcirstc.noahlab.com.hk/STC2/stc-cn.htm} released in NTCIR-12. The form of data is $(post, resp)$ pair. As the preprocessing, we pick up responses which share the same post to form $(post, {resp}_{1}, {resp}_{2}... {resp}_{n})$.

\subsection{Sentiment Analysis}
\label{senti_score}
\begin{CJK}{UTF8}{gkai} 
How to compute the sentiment of a sentence is a difficult task, which is beyond the scope of this paper. We utilize the public Hownet sentiment lexicon\footnote{http://www.keenage.com/}, which is widely used in Chinese NLP. The positive vocabulary consists of 4,530 words, such as 喜欢/love and 好/good. The negative vocabulary consists of 4,322 words, such as 憎恶/hate and 丑陋/ugly. In either of the vocabulary lists, there exist some monosyllable words, which are ambiguous and frequently used in text and so might create noisy data, such as 是/is, 能/can and 大/big in the positive vocabulary and 老/old,小/small and 黑/black in the negative vocabulary. In our experiment, we removed these six words from the lexicon.  
\end{CJK}

We calculate the sentiment score of one response sentence by using the following Equation:

\begin{equation}
\label{sent-score}
scoreSent = N_{pos} - N_{neg}
\end{equation}
where $N_{pos}$ denotes the number of words in this sentence containing in the positive vocabulary and $N_{neg}$ is the number of words in the negative vocabulary. If $scoreSent>0$, this sentence is positive; if $scoreSent<0$, it is negative; when $scoreSent=0$, we consider that this sentence doesn't contain obvious sentimental meaning.     

\subsection{Data Selection}

We first collect $(post, resp_1, resp_2)$ triples, which meet all the conditions below: 1) these two responses share the same post; 2) $resp_1$ is positive while $resp_2$ is negative; 3) the length of both responses is more than 5 tokens. We remove those responses whose length are less than 5 tokens, in order to avoid generating short and meaningless sentences. Finally, we get 278,312 such triples corresponding with 92,733 unique posts, and per post has three triples on average.  

In order to feed this data to the sequence-to-sequence neural network, we further change the form of data into $(post, resp, label)$, where sentiment label $1$ means positive while $-1$ is negative. Thus a triple $(post, resp_1, resp_2)$ is rewritten to two instances, and the final size of data is 556,624.

We randomly sample 445,330 instances ($80\%$) as training data, 55,662 ($10\%$) as validation data, and 55,662 ($10\%$) as test data. Different from the previous work, we make a “strict” split. Since a post may appear many times in different instances, we make sure that the post occurring in the training set does not appear in the test data.

\begin{figure}
\centering
\includegraphics[scale=0.30]{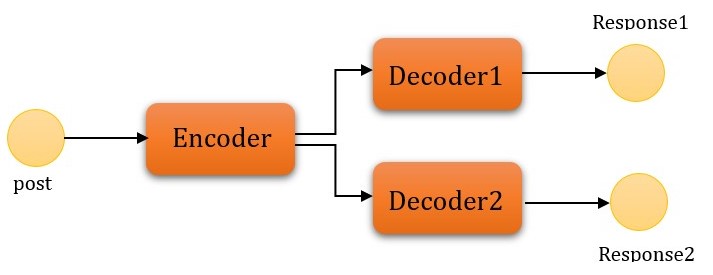}
\caption{\label{pic-framework} The framework of our dual-decoder model.}
\end{figure}

\section{Model Description}
To generate different sentiment responses, instead of adding a sentiment control mechanism on a single decoder \cite{zhou2018emotional}, we propose a dual-decoder model by connecting two sentiment decoders with one encoder, as shown in Figure \ref{pic-framework}. Our dual-decoder model enjoys the following advantages: 1) we employ two decoders respectively take responsibility of generating responses with only one particular sentiment, which makes the learning target more clear; 2) there are different low-frequent sentiment words existing in positive or negative responses separately, which might be ignored by the standard single decoder but can be efficiently captured by our dual decoders. 

\subsection{Encoder}
We adopt the encoder-decoder framework \cite{encoder-decoder2014} with attention mechanism \cite{bahdanau2015}. A post $\emph{P} = (\emph{e}_{1}, \emph{e}_{2}...\emph{e}_{n})$ is encoded into  a series of hidden representations ${h}_{p}$ via a GRU encoder:

\begin{equation}
\label{encoder-base}
\emph{h}^p_t = GRU({s}_{t-1}, \emph{h}^{p}_{t-1})
\end{equation}



\subsection{Dual Decoders}
\label{sect:PND}
Let $s^{pos}_t, s^{neg}_t$ denote the hidden state of two decoders at time \emph{t}, which can be respectively computed by:

\begin{equation}
\label{decoder-pos}
{s}_t^{pos} = GRU_{pos}({s}^{pos}_{t-1}, [{r}_{t-1}, \emph{h}^{p*}_{pos,t-1}])
\end{equation}

\begin{equation}
\label{decoder-neg}
{s}_t^{neg} = GRU_{neg}({s}^{neg}_{t-1}, [{r}_{t-1}, \emph{h}^{p*}_{neg,t-1}])
\end{equation}
where ${r}_{t-1}$ is the embedding of input word and $\emph{h}^{p*}_{pos,t-1}, \emph{h}^{p*}_{neg,t-1}$ is the response representation calculated by using the bahdanau attention mechanism.



We use ${s}_t^{pos}, {s}_t^{neg}$ to calculate the probability distribution of vocabulary $\emph{P}^{vocab}_{pos,t}, \emph{P}^{vocab}_{neg,t}$ by respectively feeding them though one linear layer and a softmax function. $\emph{loss}^{pos}, \emph{loss}^{neg}$ is the positive and negative log likelihood of probability respectively. The final loss is then defined as:

\begin{equation}
\label{PN-loss}
\emph{loss} = \alpha \emph{loss}^{pos} + \beta \emph{loss}^{neg}
\end{equation}
where $ \alpha $ and $ \beta $ are two hyper-parameters, which are used to control the relative importance of two loss functions. We set the hyper-parameters are: 

$$ (\alpha, \beta)=\left\{
\begin{aligned}
(1, 0) && if && label = 1 \\
(0, 1) && if && label = -1 
\end{aligned}
\right.
$$



\begin{table*}[h]
\centering
\small
\begin{tabular}{ccccccc}
\hline
model & sentiment accuracy &  distinct-1 & distinct-2 &  bleu-1 & bleu-2  & average\\
\hline
\hline
\multirow{1}{*} seq2seq-att & 41.2 & 76.3 & 86.3 & \textbf{14.7} & 1.22 & 35.1\\
\multirow{1}{*} seq2seq-att-sent & 75.9 & 76.1 & 84.2 & \textbf{14.7} & \textbf{1.29} & 38.9 \\
\hline
\multirow{1}{*} dual-decoder & \textbf{91.0} & \textbf{87.7} & \textbf{94.1} & 14.3 & 1.25 & \textbf{39.7}\\
\hline
\end{tabular}
\caption{\label{Experiment Result} Experimental results of different models on generating responses with sentiment.}
\end{table*}

\begin{figure*}[t]
\centering
\includegraphics[scale=0.50]{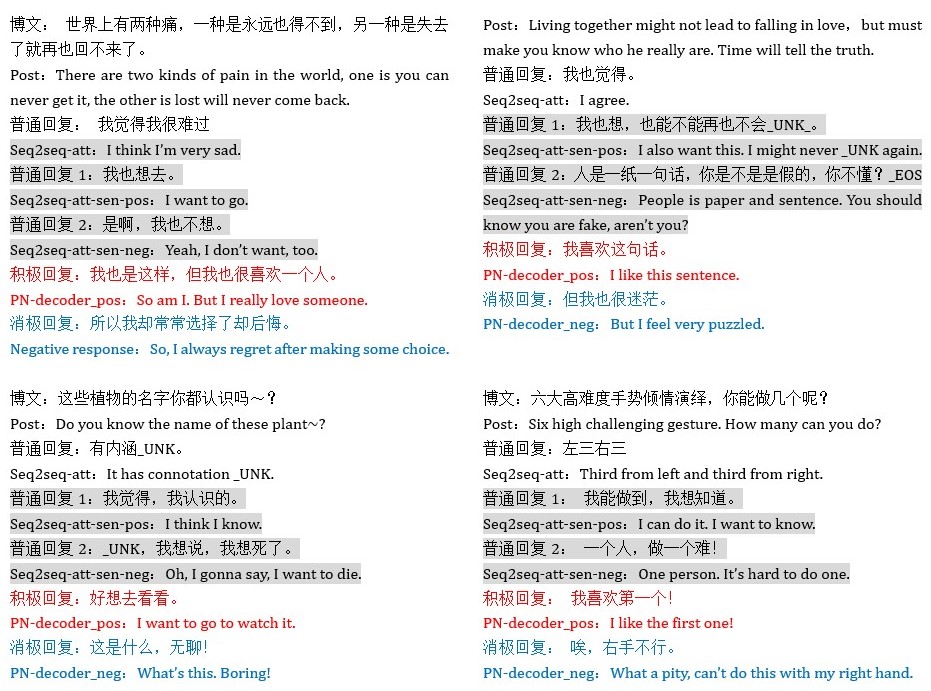}
\caption{\label{output-exp} Examples of responses generated by different models. Responses in black are generated by the baseline model, and the red colour means the positive response and blue colour means the negative response generated by our model.}
\end{figure*}

\section{Experiment}
\label{sect:experiment}

\subsection{Experiment Setup}
\label{ssec:evaluation}
We adopt the following evaluation metrics: BLEU \cite{embedding2017}, Average \cite{embedding2017}, distinct-1 \& distinct-2 \cite{distinct2016a}. In order to evaluate the performance of generating sentiment responses, we propose a new evaluation metric. For every generated response we first calculate the sentiment score according to the method described in \ref{senti_score}, and then compute sentiment accuracy that is defined as the ratio of generated responses with proper sentiment. 

In our experiment, the dimension of word embedding is set to 300. Both the encoder and decoder have 100 hidden neurons for a single network. The learning rate is fixed to 0.0002. We use Adam optimizer \cite{adam2014} to train the model. We select 30,000 most common used words respectively from posts and responses, which covers 96.2\% and 96.5\% words appearing in posts and responses. 


\subsection{Comparison Baselines}
{\bf Seq2seq-att} In this baseline model, like most of previous work, we ignore the sentiment label and treat it as a simple sequence-to-sequence generation task. 

{\bf Seq2seq-att-sent} We add sentiment information to the classical encoder-decoder framework with a single decoder, like the base model in \cite{zhou2017mojitalk}. Similar to word embedding, we use a randomly initialized vector to embed sentiment label and put it through a dense  layer to get sentiment representation $\emph{h}_s$. This model encodes post into $\{\emph{h}^p\}$ which is used latter in attention mechanism. The initial hidden state of decoder is the concatenation of encoder's last hidden state $\emph{h}^p_n$ and sentiment representation $\emph{h}_s$.

\subsection{Experimental Results}
Table \ref{Experiment Result} shows the evaluation results of our experiments. Compared with the traditional single-decoder model, our dual decoders can generate diverse responses with given specific sentiment. The three models obtain comparable results in terms of BLEU score and Average value, but we can observe a big difference in sentiment accuracy and distinct value. 

The baseline Seq2seq-att model doesn't leverage explicit sentiment information and so gets a quite low value in sentiment accuracy. The Seq2seq-att-sent model incorporates explicit sentiment information in its encoding and so improves sentiment accuracy by 34.7 points. Our dual-decoder model further improves sentiment accuracy by 25.1 points over the traditional single-decoder model. Looking at the distinct value, our model outperforms the traditional single-decoder model by nearly 10 points. We will give a detail analysis on these two aspects.  

\textbf{Diversity Analysis} To avoid repetitive sentences and to generate diverse responses is an important goal for automatic response generation system. Our model adopts two decoders separately based on positive or negative examples, and thus might capture some low frequent sentiment words. To further validate this assumption, we calculate the number of sentiment words in the responses generated by different models, which are shown in Table \ref{Senti_dis_1}. Compared with the seq2seq-att-sent model, our model uses more diverse sentiment words when generating responses. Moreover, when generating a response with a given specific sentiment, our model can use words with corresponding sentiment more correctly.

\begin{table}[h]
\centering
\small
\begin{tabular}{ccccc}
\hline
sentiment &  positive & negative &  \\
model &  words/tokens & words/tokens & Err \\
\hline
\hline
\multirow{1}{*} seq2seq-att-sent & 326 / 30,018 & 376 / 27,992 & \\
\multirow{1}{*}	--label = 1 & 312 / 295,65 & 51 / 446 & 14.0\% \\
\multirow{1}{*}	--label = -1 & 53 / 453 & 369 / 275,46 & 12.6\% \\
\hline
\multirow{1}{*} dual-decoder & 496 / 32,148 & 591 / 30,620  & \\
\multirow{1}{*}	--label = 1 & 486 / 31,830 & 44 / 533 & 8.3\% \\
\multirow{1}{*}	--label = -1 & 39 / 318 & 587 / 30,087 &  6.2\% \\
\hline
\end{tabular}
\caption{\label{Senti_dis_1} The number of positive and negative words appearing in the generated responses. The first number means how many distinct sentiment words there are while the second number counts the total number of appearing sentiment tokens.
Error rate measures how many percents of sentiment words are used incorrectly.}
\end{table}

\textbf{Case Study} Some examples generated by different models are shown in Figure \ref{output-exp}. We can see that our generated responses carry obvious sentiment meaning and show more diverse expressions.  

\section{Conclusion and Future Work}
\label{sect:conclusion}
This paper proposes a new dual-decoder framework to generate response with given sentiment, which yields significant performance gain in sentiment accuracy and word diversity. In the future, we will adopt more sophisticated neural models and feed them to our dual-decoder framework for response generation. 


\bibliography{ACL19}
\bibliographystyle{acl_natbib}

\end{document}